\documentclass[journal, twoside]{IEEEtran}
\let\labelindent\relax
\usepackage{graphicx}
\usepackage{subcaption}
\usepackage{eqlist}
\usepackage{amsfonts}
\usepackage{tabularx}
\usepackage{booktabs}
\usepackage{amssymb}
\usepackage{amsmath}
\usepackage{enumitem}
\usepackage{threeparttable}
\usepackage[vlined, ruled, linesnumbered, commentsnumbered]{algorithm2e}
\usepackage{lipsum}

\usepackage[usenames,dvipsnames]{xcolor}
\usepackage{comment}

\usepackage{multirow}

\usepackage{breqn}
\usepackage{bm}
\DeclareRobustCommand*{\ora}{\overrightarrow}

\begin{document}
\title{Four-Arm Collaboration: Two Dual-Arm Robots Work Together to Maneuver Tethered Tools}
\author{Daniel  S\'anchez$^{1}$, Weiwei Wan$^{1*}$, Keisuke Koyama$^{1}$, and Kensuke Harada$^{1,2}$
\thanks{$^{1}$Graduate School of Engineering Science, Osaka University, Japan.}%
\thanks{$^{2}$National Inst. of AIST, Japan.}%
\thanks{$^{*}$ Contact: Weiwei Wan, {\tt\small wan@sys.es.osaka-u.ac.jp}}%
}
\markboth{ARXIV Preprint. Under Review for Formal Publication, 2020}
{S\'anchez \MakeLowercase{\textit{et al.}}: Four-Arm Collaboration: Two Dual-Arm Robots Work Together to Maneuver Tethered Tools} 
\maketitle

\begin{abstract}
In this paper, we present a planner for a master dual-arm robot to manipulate tethered tools with an assistant dual-arm robot's help. The assistant robot provides assistance to the master robot by manipulating the tool cable and avoiding collisions. The provided assistance allows the master robot to perform tool placements on the robot workspace table to regrasp the tool, which would typically fail since the tool cable tension may change the tool positions. It also allows the master robot to perform tool handovers, which would normally cause entanglements or collisions with the cable and the environment without the assistance. Simulations and real-world experiments are performed to validate the proposed planner. 
\end{abstract}

\section{Introduction}

\begin{figure}[!tpb]
  \begin{center}
  \includegraphics[width=.97\linewidth]{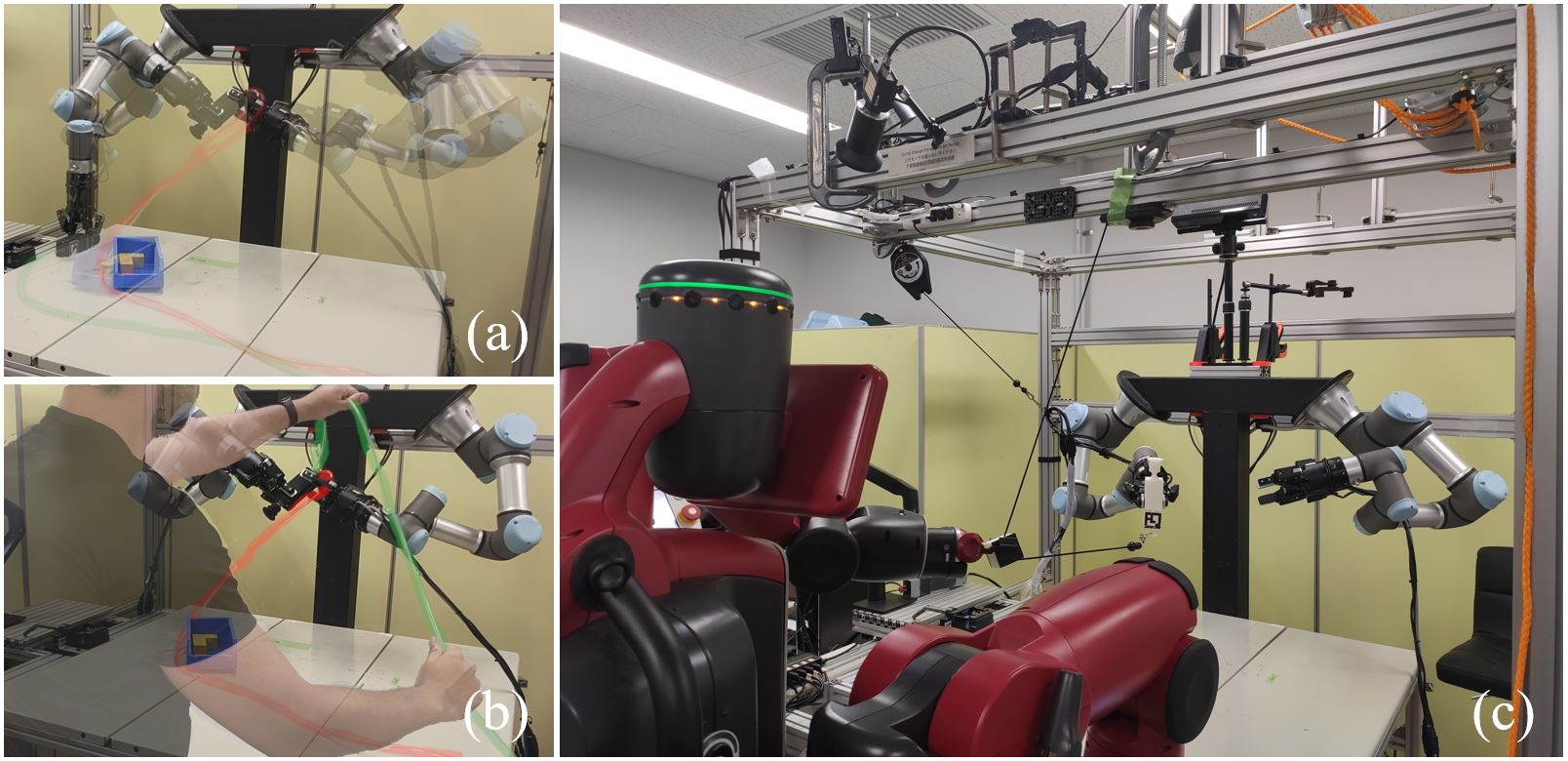}
  \caption{(a) A robot performs tethered tool manipulation by itself. The tool cable (in red) collides with objects in the workspace and gets wrongly grasped by the robot. (b) A person intervenes to remove the blocked cable, moving it to a controlled and safe state (green) to continue the robot task. (c) Inspired by the human intervention, we include an assistant robot to replace humans and help the master robot complete its task without cable accidents. The assistant robot maneuvers the cable and avoids obstacles while the tool manipulation task is being performed.}
  \label{fig:teaser}
  \end{center}
\end{figure}

\IEEEPARstart{T}{ethered} tools are widely seen in the manufacturing industry. These tools have an attached elastic cable with problematic behavior. During manipulation, cables of tethered tools will react and change their shape according to the movement of the tool, the position of the cable source, the cable tension, and the cable's inner properties. Especially when a robot autonomously manipulates a tethered tool, if the cables are unaccounted for during robotic manipulation planning, the cable can collide with objects in the robot environment, get wrongly grasped by the robot, or get entangled around the robot or the environment as seen in Fig.\ref{fig:teaser}. For these reasons, developing robust planning methods for robots to handle tethered tools are highly demanded.  

Handling tethered tools using robots can be considered as a double manipulation problem. It involves static object manipulation (tool) and indirect elastic object manipulation (tool cable). Simultaneously performing both kinds of manipulation is difficult for a single robot. Previously, we developed tethered tool manipulation planning methods for a dual-arm robot. Our methods can help avoid cable entanglement during the manipulation of tethered tools. However, they come with drawbacks such as sacrificing one robot arm to handle the tool cable \cite{sanchez2020tethered}, which prevents tool handovers, and limits the number of possible solutions for the manipulation task due to the constraints applied to the robot movement to prevent entanglements \cite{sanchez2020towards}. It remains problematic to find a manipulation planning method for tethered tools that can avoid cable entanglement and allow complex manipulation that involves handover and placement-regrasp. 

The remaining problem inspires this study. Our goal is to develop tether tool manipulation planners with special attention paid to scenarios that require re-grasping the tethered tool. Such scenarios include handing over a tethered tool and placing down a tethered tool. During handover, the tethered tool is switched from one hand to another. The cable may block feasible arm and hand motion. During placements, the tethered tool is placed down on some fixtures. The cable tension may move the tool out of position before a re-grasp is conducted. We expect our developed tool manipulation and cable maneuver planners can avoid cable problems while allowing operations like handover and placements.

Our solution's essential idea is based on an observation of what humans do when a failure is predicted in our daily experiments. When humans predict cable failures, they would play an assistant's role and actively move the cables to help avoid a robotic emergency. The detailed assistance includes lifting a cable to avoid colliding with the surrounding environment, pulling a cable to avoid entanglements with robotic bodies, stretching a cable to make it straightforward and easy to predict, etc. The human assistance role can be replaced and played by an assistant robot to perform similar autonomous manipulation, prediction, and planning. The assistant robot is expected to move accordingly with the tool-manipulating (master) robot with planned motion and modify the cable shape and trajectories to avoid collisions and entanglements. Fig.\ref{fig:teaser} illustrates the above idea with both images showing a human's intervention and an assistant robot that replaces the human's role.

To implement the assistant robot, we develop a method to plan four-arm collaborative motion considering both a tool and the related cable. The problem is complex -- It requires the simultaneous manipulation of a tool by a dual-arm master robot and manipulating the cable by a  dual-arm assistant robot. We solve the complex problem by proposing a hierarchical planning framework. Our framework firstly plans a motion sequence to manipulate the tool while disregarding the cable and then creates a motion sequence for the assistant robot to modify the cable shape and prevent collisions. We introduce cable handling gadgets to facilitate robot-robot collaboration and the control of the cable shapes. The cable handling gadgets are mechanical components or components with light mechatronic integration. They ease the prediction and simplify the constraints in planning.

In the experimental section, we carry out real-world experiments to test the proposed planner. Two cable handling gadgets and several benchmarks are used to test the generality of our solution. Furthermore, the proposed planner is also compared to other manipulation planning methods to demonstrate the benefits and observe the drawbacks. Conclusions are derived from the experimental results. 

The rest of this paper is structured as follows. Section II presents a survey on related publications. Section III shows a background of the cable handling gadgets used to ease the planning. Section IV details the steps followed by our proposed hierarchical framework for motion planning to generate a master-assistant motion sequence. Section V presents the experimental results of both simulations and real-world implementations of our framework. Finally, Section VI presents conclusions derived from the experimental results and a discussion on future works.

\section{Related work and Contributions}

This paper develops a multi-robot collaboration planner for tool manipulation while considering tethered cables. Accordingly, we review related publications in the manipulation of cable-like objects, motion planning considering tethered cables, and motion planning in the presence of multiple agents. We also summarize our difference and contributions by comparing with them.

\subsection{Manipulation of Cable-Like Objects}

Manipulation planning for deformable and cable-like objects is a difficult problem. Some early work related to this topic include a recursive learning approach to model for deformable object manipulation \cite{howard1997recursive}, using a PID controller for the manipulation of deformable objects \cite{wada2001robust}, and a visual-based approach for manipulating a rope using a dual-armed robot \cite{matsuno2001flexible}. Following them, later work further explored deformable object manipulation, some of them include a motion planner for knotting a rope \cite{yamakawa2010motion}, an implementation for robotized assembly using a wired harness in a car production line \cite{jiang2011robotized}, a motion planning algorithm for cloth folding \cite{yamakawa2011motion}, etc.

More recently, several methods have been proposed to tackle elastic object manipulation considering prediction models. For example, in \cite{duenser2018interactive} an interactive, simulation-based control methodology is proposed. It allows for user-specified deformations for elastic objects to be mapped to joint angle commands. A physics simulation engine to predict elastic object behavior is used for the manipulation planning of deformable objects in \cite{7862405}. Tactile sensors are also used to handle and manipulate cables, as presented in \cite{she2019cable}. A framework for cable shape manipulation with a dual-armed robot is discussed in \cite{zhu2018dual}. Moreover in \cite{sintov2020motion}, a motion planning strategy for elastic rod manipulation is proposed.

In this study, we do not directly model the soft deformation of cables. Instead, we employ cable handling gadgets to convert them into straight lines, thus simplify the prediction of cable shapes and planning.

\subsection{Motion Planning Considering Tethered Cables}

An anchored cable in a robot's environment can cause accidents and unforeseen entanglements with the cable. Lots of studies have proposed solutions for planning motions in the presence of tethered cables. Early work includes a path planner for a tethered robot \cite{xavier1999shortest}, a motion planning method for achieving the desired cable configuration using multiple tethered robots \cite{hert1996ties}, etc. Meanwhile, the "hyper-tether" approach was introduced. The approach is used to actively control the tether cable's tension and length as well as communication, power supply, etc., of a robot system \cite{fukushima2000new}\cite{fukushima2001development}.

The early studies inspired the innovation of several tethered robotic systems and the study of tethered robot dynamics and path planning. For example, in \cite{Capua2009Motion}, a study on the kinematics, statics, motion planning, and the design of an under-constrained cable-suspended robot is shown. An experimental comparison of motion between robot systems with active and passive tether cables is shown in \cite{Yang2009Experimental}. In \cite{Kim2014Path}, a path planning algorithm for tethered robots in cluttered environments is presented. 

More recent literature also studies the topic substantially. For example, a hydrodynamics and control model to simulate a tethered underwater robot system is presented in \cite{wu2018integrated}. A motion planning algorithm for multiple tethered planar mobile robots, emphasizing entanglement avoidance, is proposed in \cite{zhang2019planning}. An approach for cooperative manipulation of a cable-suspended load with two aerial robots is presented in \cite{Tognon2018Aerial}.

Our proposed approach for tethered tool manipulation borrows the concept of controlling the tether tension from the hyper-tether approach. In our method, we plan robot motion while taking the advantages of cable handling gadgets. The gadgets provide tension to the cables.

\subsection{Multi-agent Motion Planning and Collaboration}
The key point of motion planning for multiple robots or agents is coordinating the robots' movements and goals. Seminal work on multi-robot motion planning has explored several specific problems of multi-robot planning. For example, in \cite{736775}, a motion planner for multiple robots with different goals is presented. In \cite{chang1990collision}, a collision-free method for two articulated robot arms is shown. An evolutionary approach for the collision-free motion planning of multiple robot arms is presented in \cite{rana1995evolutionary}. 

Following up, an algorithm for diminishing operational times while using two robot arms is presented in \cite{kurosu2017simultaneous}. In \cite{dobson2017scalable}, a scalable, sampling-based planner for coupled multi-robot motion planning problems is presented. An integrated motion and task planning methodology for multiple robot arm systems is proposed in \cite{umay2019integrated}. In \cite{wang2018trust}, an approach for multi-robot motion planning with a human in the loop is shown. In \cite{wang2016kinematic}, a decentralized algorithm that coordinates the forces of a group of robots for cooperative manipulation is presented.

Researchers have also focused their efforts on improving industrial robot-human collaboration\cite{tsarouchi2017human}\cite{maurtua2017human}\cite{robla2017working}. Collaborative robot-robot interaction also draws lots of the community's attention\cite{hadidi2018}\cite{asfour2018armar}. These studies helped advance robotic collaboration in industrial scenarios.

In this paper, we employ two dual-arm robots to conduct four-arm coordination, and plan the motion for four robotic manipulators. Our planning is hierarchical -- First, we plan the master robot's motion and use it to determine the tool's motion. Then, we employ the motion of the tool to determine the motion of a cable handling gadget. Finally, the motion of the assistant robot is planned to move the cable handling gadget accordingly.

\subsection{Contributions}
The previously mentioned work offers solutions for manipulating cable-like objects and multiple robot collaboration and motion planning. However, few discussions were made to the influence of a tethered cable during object regrasping using handover and placements. The advantages of four-arm collaboration in complex manipulation tasks involving re-grasping are also less unexplored.

In our previous work, we presented planning solutions to avoid cable entanglements by constraining the cable entanglement  \cite{sanchez2020towards} and by directly modifying the cable trajectory with a cable slider using a dual-armed robot \cite{sanchez2020tethered}. However, there remain limitations. In the first case, the robot cannot perform tool placements on the workspace table due to the cable tension, limiting and restricting the number of possible solutions to any given manipulation task. In the second case, a free robotic hand can directly control the cable shape, which forbids the robot from performing tool handover. Without handovers, the robot can only place the tool at positions where its tool-manipulating hand can reach. It cannot switch the tool between hands, which limits the effective work range. In both implementations, the robot movements are constrained by the cable.

The solution proposed in this new manuscript aims at addressing the mentioned limitations by delegating the cable entanglement avoidance task to an assistant robot. The assistant robot moves simultaneously with the master robot that handles the tool. It changes the shape and state of the tool cable to prevent the master robot from getting entangled. The solution gives the master robot an unconstrained set of manipulation tasks/goals, which allows the master robot to execute both handovers and placements of the tool to complete its task. Mechanical cable handling gadgets are used to straighten the cable shape and facilitate cable motion prediction during motion planning.  

\section{Mechanical Cable Handling Gadgets}
\begin{figure}[!htbp]
  \begin{minipage}[c]{0.615\linewidth}
    \includegraphics[width=\textwidth]{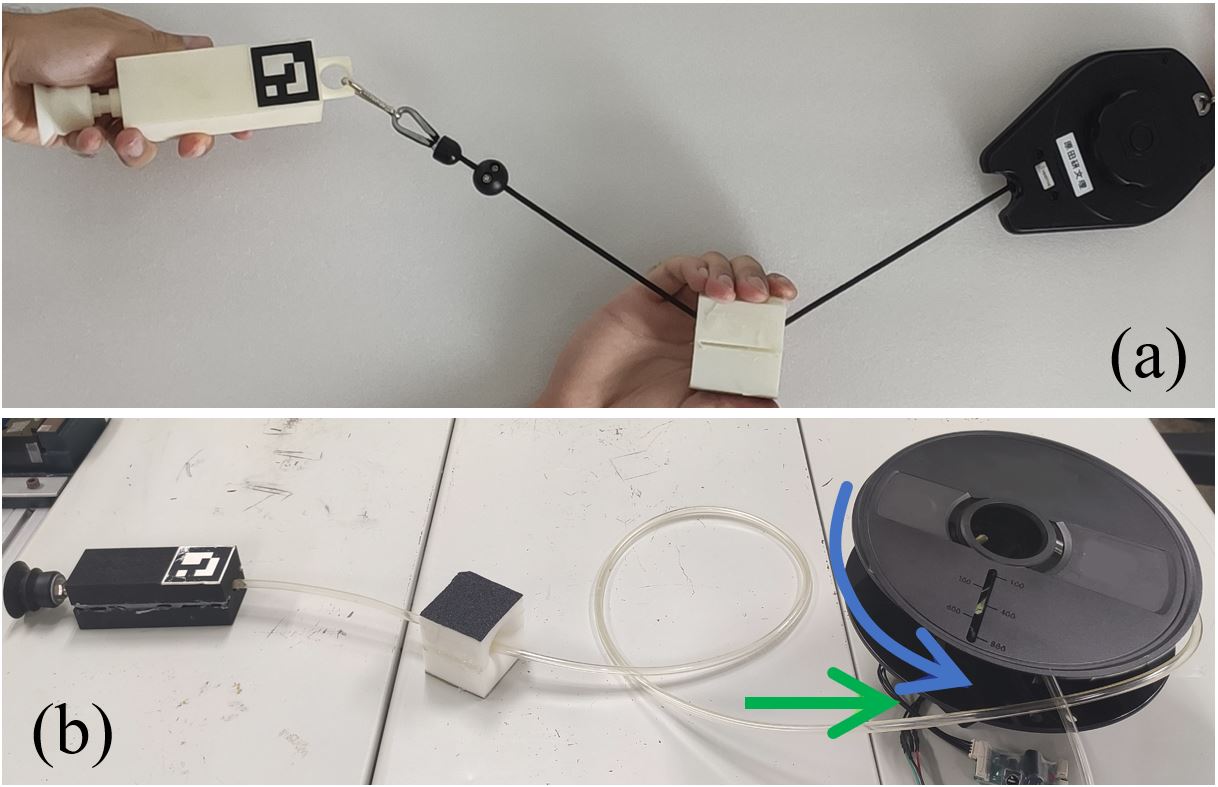}
  \end{minipage}\hfill
  \begin{minipage}[c]{0.36\linewidth}
    \caption{Two kinds of cable handling gadgets. (a) A tool balancer. (b) A motorized pulley. Long objects: The tools. Cubes: The sliders. Arrows: Blue-Rotation; Green-Cable motion.} 
    \label{fig:gadgets}
  \end{minipage}
\end{figure}

Two kinds of mechanical cable handling gadgets are used to straighten the cables and ease the planning -- A tool balancer and a motorized pulley, as seen in Fig.\ref{fig:gadgets}.

Tool balancers are purely mechanical. They connect a tool to an anchor point in the robot workspace. A tool balancer is made of a torsional springs mechanism that offers a string with constant tension. When no force is applied, the string is retracted, so does the tool. When the string is pulled, the tension generated by the torsional spring straightens the string. The string tension keeps the cable shape as a straight line connecting the tool's end to the tool balancer. In most working scenarios, a tool balancer is hanged at the top of a workspace. The anchor point where a tool balancer is attached to is known. Together with the tool balancer, a cable slider, i.e., the white cube in Fig.\ref{fig:gadgets}(a) is attached to the string to allow further segmenting the able into two halves. A robot may pull the string by grasping the slider, instead of directly touching the string. The cable slider is simply a solid block with a throughout hole for the string to pass.

The tool balancer is good at controlling the tension. However, it is difficult to replace the string using an arbitrary cable, e.g., a power supply cable, a vacuum tube, etc. Thus we prepare a second mechanical cable handling gadget, where a cable is winded around a motorized pulley. The pulley helps to retract, extend, and straighten the cable. The gadget is shown in Fig.\ref{fig:gadgets}(b). To use the gadget, one must estimate the distance the tool cable is pulled. Once the distance surpasses a given threshold, a control signal must be sent to the pulley motor to retract or extend the cable. The blue arrow in Fig.\ref{fig:gadgets}(b) indicates the rotation direction of the pulley. The green arrow indicates the resulted cable motion. Like the tool balancer, the motorized pulley simplifies a cable into straight lines and facilitates manipulation planning. A cable slider is attached to the cable to allow further bending the able into two line segments.

\section{Assisted Manipulation Planning}
Our tethered tool manipulation planning approach consists of two motion sequences for two robots that work in collaboration. The first motion sequence is used by the master robot to manipulate a tool, which is called the Task Motion Sequence (TMS) for convenience. The sequence is computed using our previously proposed single-arm manipulation planner \cite{wan2016developing}. The second motion sequence is called the Assistant Motion Sequence (AMS). Like its name, the second sequence is generated to respond to the TMS and the expected cable trajectories and shape. We use the cable handling gadgets to place the tool cable under tension and accurately predict the cable's shape and trajectory. The cable shape depends on the slider position. It is divided by the slider into two straight line segments in space. One of the segments connects the tool to the slider. The other one connects the slider to the source of the cable. Collision avoidance is performed by preventing overlap between robots and the segments. Following the AMS, the assistant robot is expected to move the cable slider to change the cable shape and avoid collisions. The cable slider trajectories are computed in accordance with the expected tool trajectories generated by the TMS. For every tool state in a tool trajectory, an IK-feasible and collision-free pose for the cable slider and the assistant robot is computed. Both the TMS and AMS are validated in simulation before being sent to the real-world counterparts.

\subsection{Task Motion Sequence (TMS) Planning}
The TMS is generated using our previous planner \cite{wan2016developing}. To generate the TMS, our planner searches a pre-built graph of object-specific grasps to connect compatible pick-up (in the object starting/original pose) and place-down (for the desired object goal pose) poses of an object. If necessary, the planner can use object placements (placing the object in the workspace table to perform a regrasp) or handovers (switching the object from one hand to another) and generate a multi-hop path between the starting tool grasp pose and the goal tool grasp pose.

\subsection{Cable Shape Prediction}
When our planner generates the TMS, it also predicts the tool poses and trajectories during manipulation. Every tool pose in the trajectory is directly associated with the TMS and the master robot's grasp pose. Each tool pose can be used to estimate the corresponding tool cable state and shape. If the cable is under tension, the cable shape can be estimated as a straight line connecting a source node of cable (the tool balancer or place where the cable comes from) and a sink node, where the cable attaches (a specific point on the tool), as seen in Fig.\ref{fig:gadgets}. Furthermore, by adding a cable slider, it is possible to modify the cable's shape -- The slider introduces a pair of sink-source nodes for the cable and divides it into two separate segments. We use two vectors to denominate the cable segments at every discretized robot pose of the robot motion sequence. The first vector $\overrightarrow{\mu_{ts}}^{l}$ is the vector of the cable segment connecting the tool to the bottom of the cable slider for the $l$-th pose of the master robot TMS. It can be computed using equation \eqref{eq:tool to slider segment}: 
\begin{equation}\label{eq:tool to slider segment} 
\overrightarrow{\mu_{ts}}^{l} = s^{l}- t^{l},
\end{equation} 
where $s^{l}$ is the $l$-th position of the cable slider bottom and ${t}^{l}$ is the $l$-th position in which the cable attaches to the tool. Both of the positions are defined in the world reference frame. 

The second vector $\overrightarrow{\mu_{sp}}^{l}$ is the cable segment that connects the top of the slider to the source of the cable (pulley or balancer). This vector is computed using equation \eqref{eq:slider to balancer segment}:
\begin{equation}\label{eq:slider to balancer segment} 
\overrightarrow{\mu_{sp}}^{l} = \textit{h} - s^{l} + \overrightarrow{b}^{l},
\end{equation} where $\textit{h}$ is the known position of the cable source location in the world frame of reference and $\overrightarrow{b}^{l}$ is the offset between the bottom position of the slider $s^{l}$ and the top. It is important to notice that $s^{l}$ directly affects both vectors. We use said principle to move the slider position across the robot workspace to control the vectors and hence the cable shape and its trajectory. 
 
 \begin{figure}
  \begin{minipage}[c]{0.5\linewidth}
    \includegraphics[width=\textwidth]{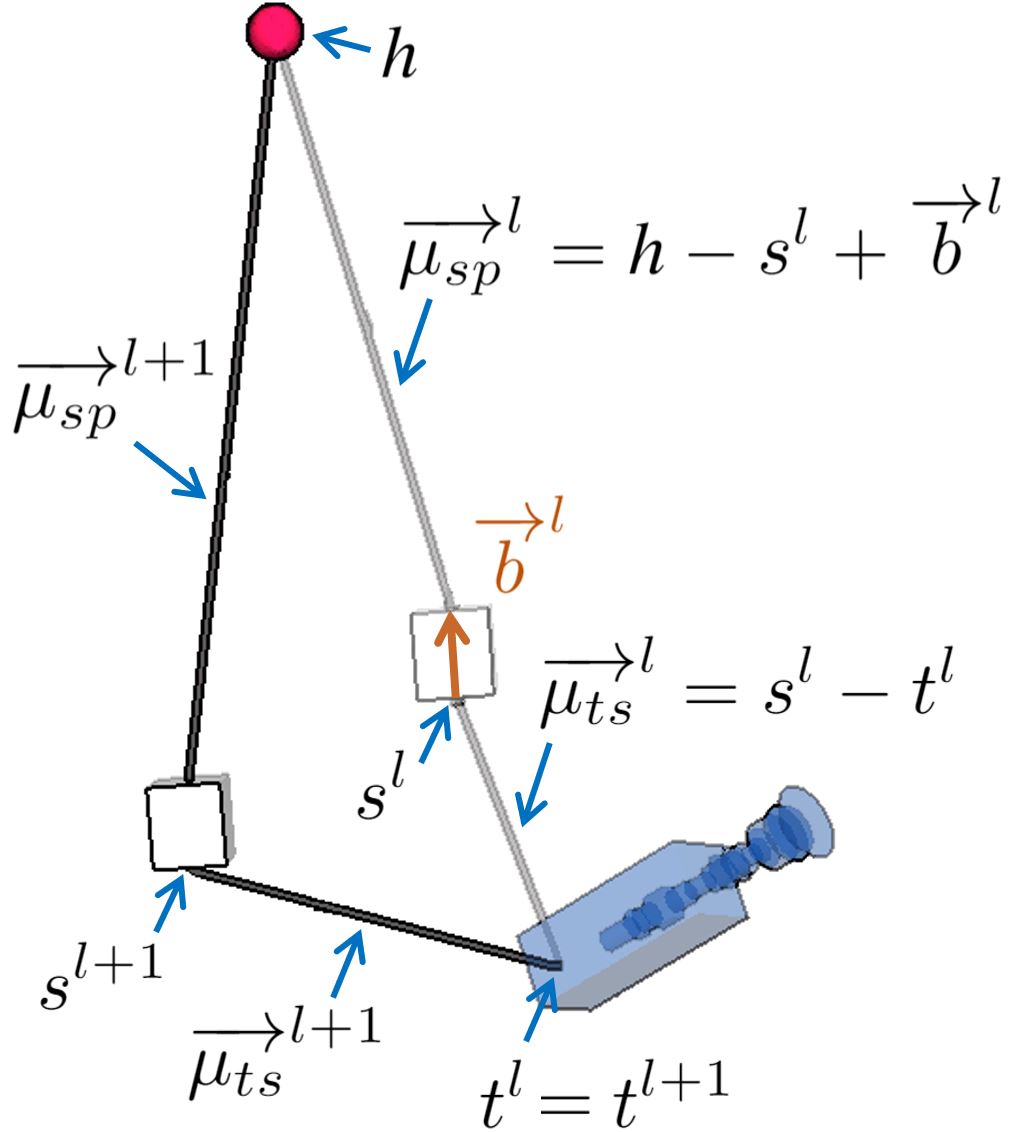}
  \end{minipage}\hfill
  \begin{minipage}[c]{0.47\linewidth}
    \caption{When under tension a cable is divided into two line segments $\ora{\mu_{ts}}^{l}$ and $\ora{\mu_{sp}}^{l}$. They are determined by the slider position $s_{i}^{l}$ and the position of the cable source $h$. By changing the slider position, the assistant robot controls the two line segments and the cable trajectories to avoid entanglements and collisions.} 
    \label{fig:Ancgleacc3D}
  \end{minipage}
\end{figure}

\subsection{Planning the Trajectory of the Slider} 
Our proposed planner generates a desired trajectory for the cable slide by considering the tool trajectory determined by the TMS. More specifically, for every discrete tool pose in the tool trajectory, the planner selects a pose from a list of poses for the cable slider. The slider poses are constrained following these rules: 
 
 \begin{itemize}
  \item  To avoid cable entanglements, the cable segments must not be in collision with the robots or the environment.
  \item  To avoid excess strain on the tool and robots, the distance between the tool and the slider must stay within a certain margin.
  \item  The first cable segment must have less than 90 degrees of bending in the tool's local frame. Likewise, the second cable segment must not bend more than 90 degrees in the slider's local frame.
\end{itemize}

For convenience, we represent the tool's local frame as $\Sigma_T$, and align its $\overrightarrow{z}$ axis with the normal of the tool surface that attaches to the cable. Its origin position is also set to the position where the cable attaches to the tool. By placing the tool frame of reference in the said pose, we create a relationship between the tool's $\overrightarrow{z}$ axis and the tool cable. The $\overrightarrow{\mu_{ts}}^{l}$ segment of the cable originates at the origin of $\Sigma_T$. The larger the angular difference between $\overrightarrow{\mu_{ts}}^{l}$ and $\Sigma_T$'s $\overrightarrow{z}$ axis, the more the cable is considered to be bent. A planner can prevent excess cable bending by controlling the angular difference. 

\begin{figure}[!htbp]
  \begin{center}
  \includegraphics[width=.95\linewidth]{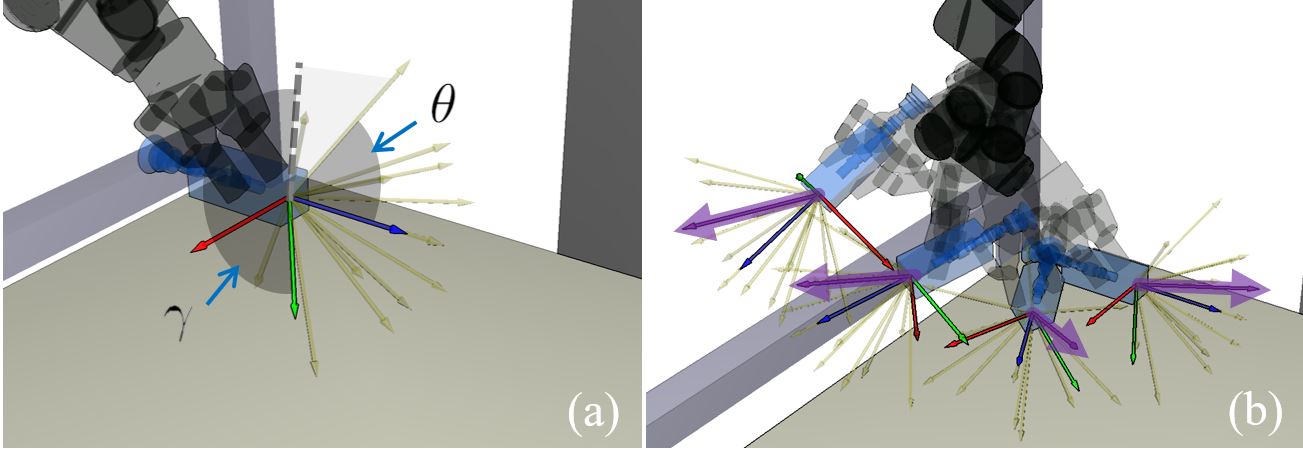}
  \caption{(a) The master robot holds a tool in the simulation environment. Our planner generates candidate positions to place a cable slider and control the tool cables. For every tool pose computed for the TMS, the planner samples candidate positions for the cable slider. The positions are defined by the tip of the yellow vectors shown in this subfigure. These vectors are sampled using the tool reference frame represented by the red ($\ora{x}$), green ($\ora{y}$) and blue ($\ora{z}$) axes, the angle from the $\ora{z}$ axis ($\theta$ in the subfigure), and the angle between the yellow vector projection on the $\ora{x}$-$\ora{y}$ plane and the $\ora{y}$ axis ($\gamma$ in the subfigure). (b) A chosen candidate slider position $s^{l} =\ora{\lambda_i}^{l} + t^{l}$ (purple vector tip) for each of the tool pose in a tool trajectory.}
  \label{fig:multicandidate}
  \end{center}
\end{figure}

For every $l$-th discrete tool pose in a tool trajectory, our planner samples a list of candidate poses that meet the aforementioned rules for the slider. The sampling is performed using two angles $\theta$ and $\gamma$. We represent a sampled candidate pose using a position ${s_{i}}^{l} = \overrightarrow{{\lambda}_{i}}^{l} + {t}^{l}$. $t^{l}$ is the position of the tool reference frame, as was seen in Fig.\ref{fig:Ancgleacc3D}. The vector $\overrightarrow{{\lambda}_{i}}^{l}$ is computed using the angles $\theta$ and $\gamma$. $\theta$ is defined as the angle between $\overrightarrow{{\lambda}_{i}}^{l}$ and $\Sigma_T$'s $\overrightarrow{z}$ axis. $\gamma$ is defined as the rotation angle of $\overrightarrow{{\lambda}_{i}}^{l}$ around $\Sigma_T$'s $\overrightarrow{z}$ axis. $\overrightarrow{{\lambda}_{i}}^{l}$ is computed using
\begin{equation}\label{eq:candidate_position}
\overrightarrow{{\lambda}_{i}}^{l} = \omega R^{T}(l)\cdot(\sin{\theta}\cos{\gamma}, \sin{\theta}\sin{\gamma}, \cos{\theta}),
\end{equation}
where $R^{T}(l)$ is the rotation matrix of $\Sigma_T$ for the $l$-th element of the TMS. The planner computes a list of $\overrightarrow{{\lambda}_{i}}^{l}$ by sampling $\theta$ and $\gamma$, i.e. $\theta \in \{30 ^{\circ},60 ^{\circ} \}$ and $\gamma \in \{0^{\circ},60^{\circ},120^{\circ},180^{\circ},240^{\circ},300^{\circ}\}$. The module parameter $\omega$ gives the vector a magnitude of 325 $mm$. Besides, we define a special base case, $\overrightarrow{{\lambda}_{0}}^{l}$, using $\theta = 0^{\circ}$ and $\gamma = 0^{\circ}$. $\overrightarrow{{\lambda}_{0}}^{l}$ essentially equals to a vector in $\Sigma_T$'s $\overrightarrow{z}$ axis, for a total of 13 different $\overrightarrow{{\lambda}_{i}}^{l}$ vectors. Fig.\ref{fig:multicandidate} illustrates the various symbols and several examples of $\overrightarrow{{\lambda}_{i}}^{l}$ at different $l$-th poses.

\subsection{Assisted Motion Sequence (AMS) Planning}
\begin{figure}[!htpb]
  \begin{center}
  \includegraphics[width=.95\linewidth]{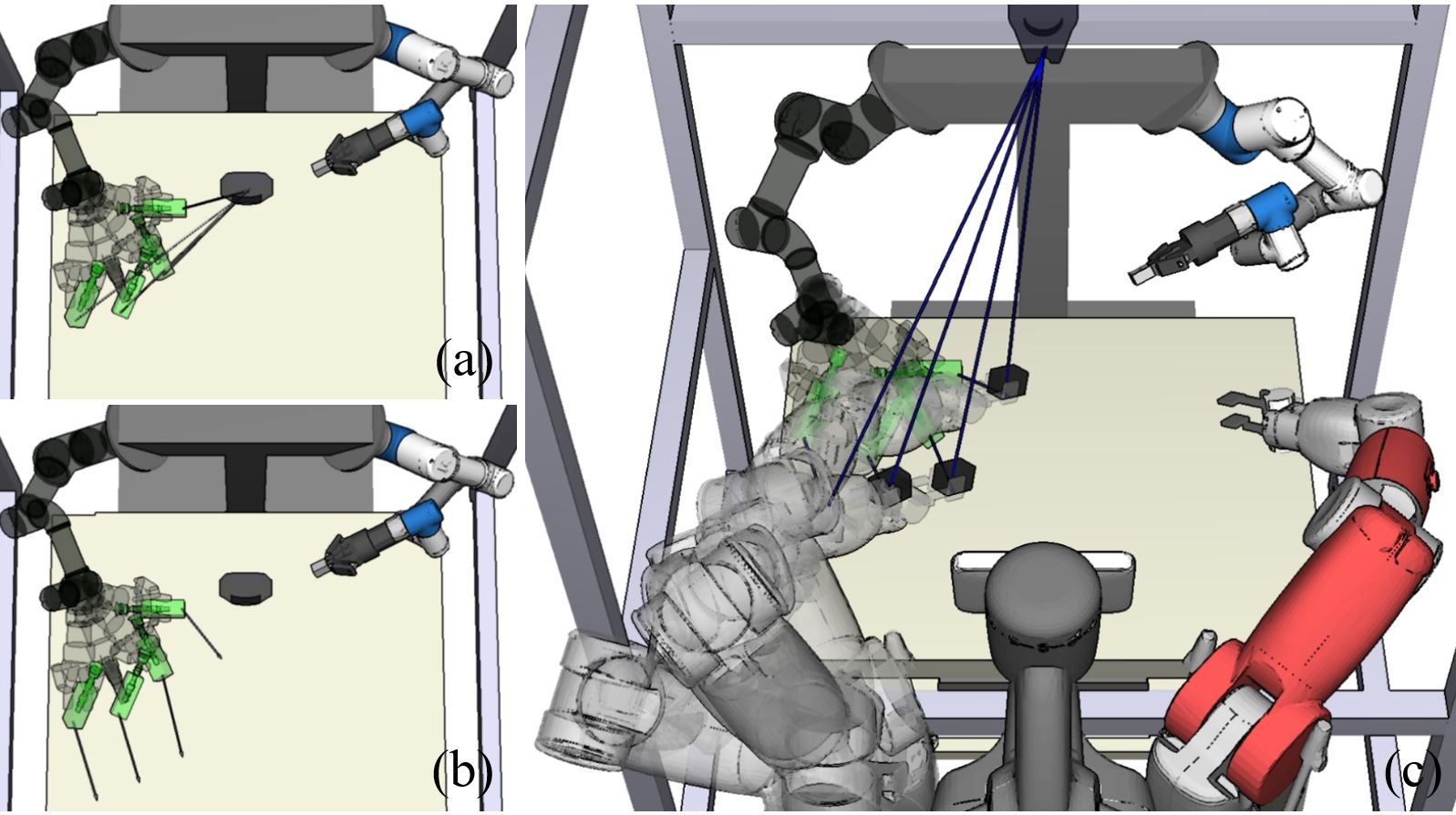}
  \caption{The proposed planner generates the AMS in our simulation environment. (a) The planned TMS. (b) For each tool pose in the TMS, our planner samples slider poses and select a sampled candidate considering line segments divided by the slider. (c) The assistant robot selects a grasp and solves the IK for holding the slider using the selected grasp.} 
  \label{fig:simulation_teaser}
  \end{center}
\end{figure}

Our planner generates a series of corresponding assistant robot poses to place the cable slider in desired positions following the elements in the TMS. For a given $l$-th tool pose $(t^{l}, R^T(l))$ of the TMS, the planner uses grasp reasoning and Inverse Kinematics (IK) to place the assistant robot slider-holding end-effector at a given position chosen from the candidate list \{${s}^{l}_{i}$\}. Alg.\ref{alg:Assistant robot pose computation} shows the process to compute the $l$-th assistant robot IK pose $IK^{l}$. The algorithm uses the \textit{nexti} and the last found vector id ${i}_{last}$ to determine the next vector id and uses the \textit{checkConstraints} function to compute IK-feasibility of the assistant robot to place the slider in the next position ${s}_{i}^{l}$. The \textit{checkConstraints} function also finds the divided line segment vectors by using \eqref{eq:tool to slider segment} and \eqref{eq:slider to balancer segment}, and checks their collisions with robot bodies and environments. If a IK-feasible and collision-free result is found, the search is finished. Otherwise, the slider pose ${s}_{i}^{l}$ is re-computed by continuously iterating to the next id. The next id $i$ is determined by selecting the $\overrightarrow{\lambda_{i}}^{l}$ candidate value that is the closest to $\overrightarrow{\lambda_{{i}_{last}}}^{l-1}$. The selection follows the sequence $i={i}_{last}$, $i={i}_{last}$+$1$, $i={i}_{last}$-$1$, $i={i}_{last}$+$2$, $i= {i}_{last}$-$2$, \dots until $i$ goes beyond the sampled size. Here, we always try $i={i}_{last}$ first to keep the robot pose as much as possible. SLERP Interpolation is performed to connect the last known feasible solution to the newly found feasible pose $IK^{l}$. The whole process is repeated for the computation of all robot poses until a motion sequence is found or a failure is reported. Fig. \ref{fig:simulation_teaser} shows an example of the generation process for the AMS in our simulation environment.

\begin{algorithm}[tbp]
\SetKwFunction{length}{length}
\SetKwFunction{nexti}{nexti}
\SetKwFunction{append}{append}
\SetKwFunction{checkConstraints}{checkConstraints}
\SetKwFunction{isReusable}{isReusable}
\SetKwFunction{reasonGrasp}{reasonGrasp}
\KwResult{AMS computation.}
$l\leftarrow 0$;
$i$$\leftarrow$${i}_{last}$$\leftarrow$0;
\textnormal{AMS}$\leftarrow$ $\emptyset$\;
\While{ \textnormal{g}$\leftarrow$\reasonGrasp{$\mathcal{G}$} } {
    \While{ l $<$ \length{$\textnormal{TMS}$} } {
        $isfound$ $\leftarrow$ \textnormal{False}\;
        \While{ i $\leftarrow$ \nexti{$i_{last}$, $i$} }{
            $s^{l}_{i}$ $\leftarrow$ $\overrightarrow{\lambda}^{l}_{i} + t^{l}$\;
            $IK^{l}$ $\leftarrow$ \checkConstraints{$s^{l}_{i}$,$t^{l}$,$h$,$\textnormal{g}$}\;
            \If{$IK^{l}$}{ 
                ${i}_{last}$ $\leftarrow$ i; i $\leftarrow$ 0; $isfound$ $\leftarrow$ \textnormal{True}\;
                \textnormal{AMS}.\append{$IK^{l}$}\;
                \textnormal{break}\;
            }
        }
        \If{\textnormal{not} $isfound$}{
            \If {\isReusable{$IK^{l-1}$, $t^{l}$, $h$, $\textnormal{g}$}} { 
                \textnormal{AMS}.\append{$IK^{l-1}$}\;
            }
            \lElse {\textbf{goto} line 2}
        }
        $l$ $\leftarrow$ $l + 1$\;
    }
    \textnormal{return} \textnormal{AMS}\;
}
\textnormal{return} $\emptyset$\;
\caption{Planning an AMS Trajectory}
\label{alg:Assistant robot pose computation}
\end{algorithm}

\subsection{Special Policies} 
\subsubsection{Cable tension during tool placement} In the cases where the robots use the tool balancer to straighten the tool cable, the tension on the cable can move the tool out of position if it is not firmly grasped. The problem disables motion like placing down the tool on a table for regrasp. To allow the master robot to perform placements and regrasp, we plan the assistant robot to use its free hand to grab and keep the cable in position while the master robot is not holding the tool. Since the tool cable forms a straight line between the top of the cable slider and the tool balancer, the planner can use IK to solve the assistant robot pose by considering its free hand at any given grasping point along the line formed by the cable. RRT is used to generate a motion sequence for grabbing and pulling the cable. Once the master robot re-grasps the tool, the assistant releases the cable.

\subsubsection{Cable over-extension} The motorized pulley allows automatic winding or releasing the tool cable when the master robot moves the cable in excess. The planer determines an excessive cable length by measuring the distance between the cable source position and the tool position during manipulation. If the difference between the measured distance and the outside cable length surpasses a given threshold, the planner activates the pulley motor to straighten the cable before continuing the robot motion sequences. 

\section{Experiments}\label{sec:experiment}
Both simulations and experiments are performed to validate our proposed planner. The simulation is based on the tool balancer gadget. In the simulation, we compare our planners with another three methods to study its performance. In the real world experiments, we validate and compare the results of using both the two cable handling gadgets.  

\subsection{Simulations and Comparisons with Other Methods}

The proposed planner, together with three previously developed methods, are implemented and compared using simulation. The three previous methods include: (1) A single TMS planner \cite{wan2016developing}, in which the cable is not considered; (2) A Cable Maneuvering motion Sequence (CMS) planner, which was previously developed for tethered tool manipulation using two collaborative arms \cite{sanchez2020tethered}; (3) Constrained Sequence with Handovers (CSH) \cite{sanchez2020towards}, which predicts the cable shape by assuming it to be a straight line and use motion constraints to avoid entanglements.
The proposed planner, the CMS, and the CSH assume using a tool balancer gadget to handle the cable. 

 \begin{figure}[!htbp]
  \begin{center}
  \includegraphics[width=.95\linewidth]{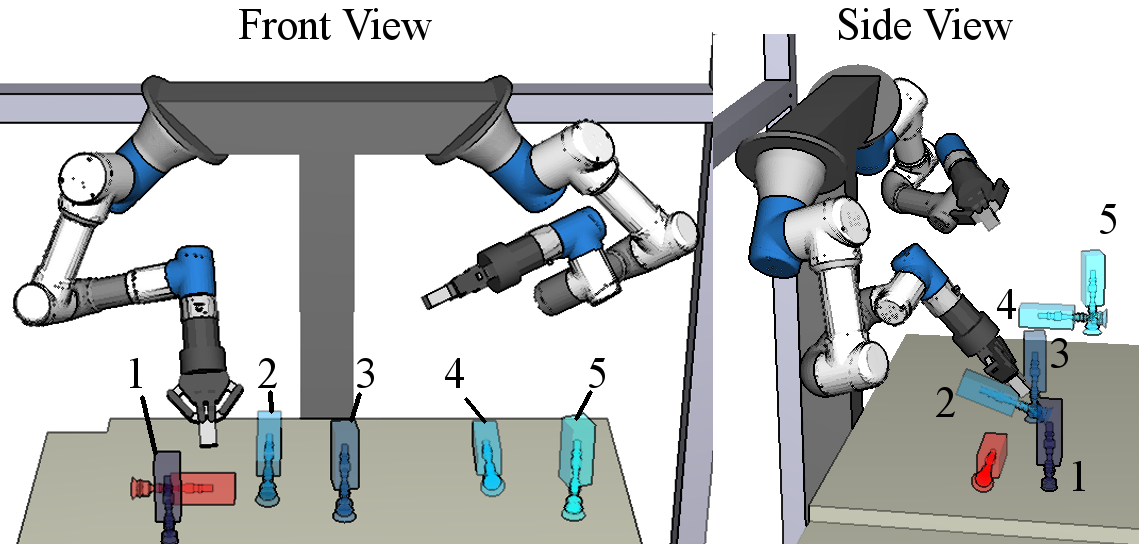}
  \caption{The starting (red) and goal (shades of blue) poses for the benchmark. Every benchmark considers the same tool starting pose and randomly selects three goal poses to place the tool during a single manipulation planning motion sequence.} 
  \label{fig:benchmarks}
  \end{center}
\end{figure} 

Several benchmarks are proposed to compare planner performance. Each benchmark consists of one tool starting position and three goal positions in which the master robot must place the tool. The starting and tool goal positions for each benchmark are chosen from a list of five different tool goal poses and a single starting tool pose, as seen in Fig. \ref{fig:benchmarks}. The tests were run on a computing system with an Intel Core i9-9900 K CPU (3.6 GHz clock) and a 32 G memory at 3600 MHz (DDR4). 

Table \ref{Tab: Benchmarks} shows the results of our simulations. The table showcases the average planning times (for five simulations per benchmark) and the success or failure to compute a motion sequence to complete the task without cable collisions.

\begin{table}[!htbp]
  \begin{center}
    \renewcommand\arraystretch{1.2}
    \caption{The angle accumulations of the benchmarks \label{Tab: Benchmarks}}
    \resizebox{0.97\linewidth}{!}{
    \begin{threeparttable}
      \begin{tabular}{ccllll}
        \toprule
       Task ID & Goals & TMS+AMS & TMS & CMS & CSH\\
        \midrule
        i & 1, 4, 3 & $\bigcirc$~$[56.70]$ & $\Delta$~$[44.32]$ & $\times$~$[-]$ & $\bigcirc$~$[90.52]$\\
        \midrule
        ii & 3, 2, 5 & $\bigcirc$~$[58.36]$ & $\Delta$~$[39.53]$ & $\times$~$[-]$ & $\bigcirc$~$[100.23]$\\
        \midrule
        iii & 2, 1, 3 & $\bigcirc$~$[48.39]$ & $\Delta$~$[45.77]$ & $\bigcirc$~$[55.34]$ & $\bigcirc$~$[100.36]$\\
        \midrule
        iv & 4, 2, 1 & $\bigcirc$~$[60.55]$ & $\Delta$~$[38.92]$ & $\times$~$[-]$ & $\times$~$[-]$\\
        \midrule
        v & 5, 1, 2 & $\bigcirc$~$[63.85]$ & $\Delta$~$[43.89]$ & $\times$~$[-]$ & $\times$~$[-]$\\
        \bottomrule
      \end{tabular}
      \begin{tablenotes}
      \item[*] Meanings of abbreviations: TMS: Tool manipulation sequence. AMS: Assistant manipulation sequence. CMS: Cable maneuvering motion sequence. CSH: Constrained sequence with handovers. $\bigcirc$: Success. $\Delta$: Succeeded in planning with cable collisions. $\times$: Failed to find a motion sequence to solve the benchmark task. The average computation times (for five iterations) are shown in seconds within brackets. 
      \end{tablenotes}
    \end{threeparttable}
    }
\end{center}
\end{table}

It is important to notice that the tool goal poses at the master robot left-hand-side are only reachable by its left hand. They are outside the workspace of the master robot's right arm. The master robot must use its right arm to initially pick up the tool from its starting position. Subsequently, tool handover is necessary to transfer the object from the robot's right hand to its left hand. The Cable manipulation (CMS) method is incompatible with handovers (since the robot must always hold the cable with one of its hands), limiting its success rate. On the other hand, the constrained motion sequences (CSH) failed for benchmarks 4 and 5 since no solution was found within the time limit of 120 seconds. In all cases, a TMS-only motion sequence was found to solve the task. However, some of the TMS solutions require the master robot to place the tool on the table and regrasp it, given the cable tension yielded by the tool balancer, said solutions are not valid. Further more, the TMS-only sequences caused the tool cable to collide with the robot at several points in the TMS. The TMS+AMS planner successfully created motion sequences that avoid the previously mentioned problems.

\subsection{Real-world Experiments}
Real-world experiments are performed to examine the adaptation of our planner to the two cable handling gadgets. The different gadgets allow handling different cable types.

\subsubsection{Results using the tool balancer} 
\begin{figure}[!htpb]
  \begin{center}
  \includegraphics[width=.95\linewidth]{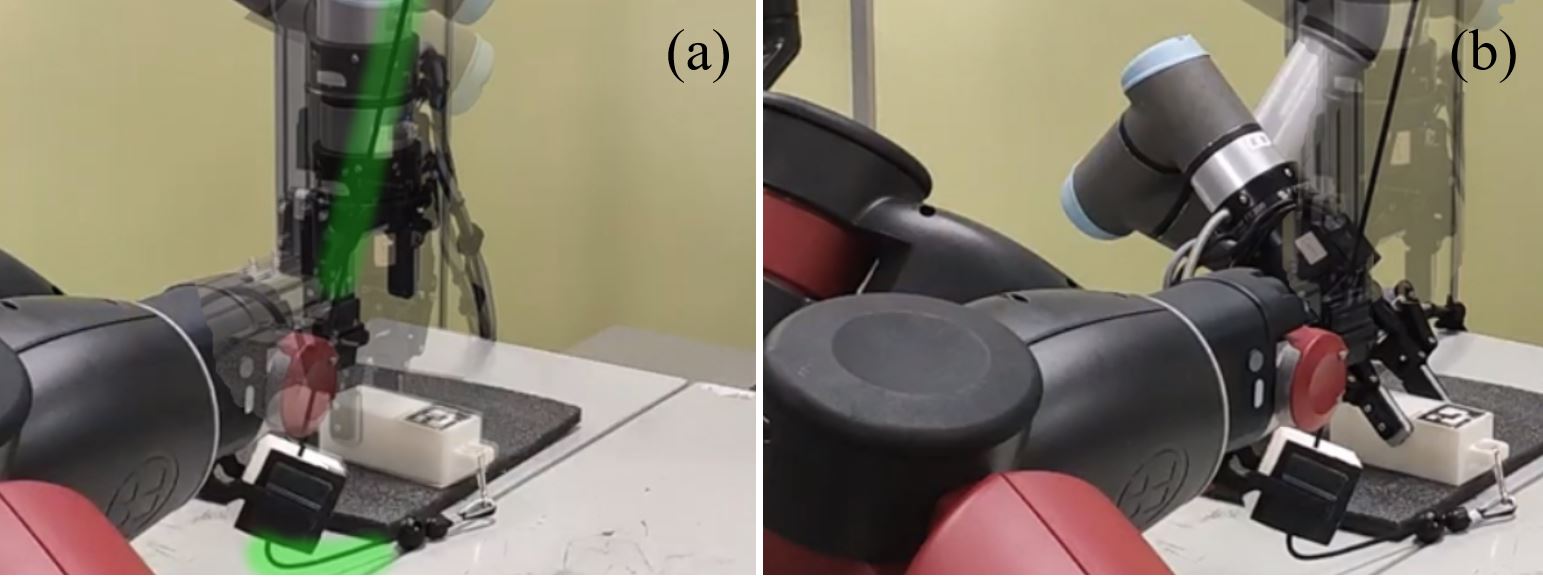}
  \caption{(a) The assistant robot holds the tool cable while the master robot releases the tool. (b) The master robot changes its grasping pose to re-grasp the tool and continue the manipulation sequence. The cable tension is suppressed by the assistant robot's cable holding hand.} 
  \label{fig:cablehold}
  \end{center}
\end{figure}

For the real-world experiments with the tool balancer, we tested the same benchmarks of Table \ref{Tab: Benchmarks}. To complete the benchmarks, the robot must perform tool placements and handovers. For the tool placements, the tool cable's constant tension would pull the tool out of position when the tool is released on the table. With our planner, the assistant robot can aid by pulling and holding the tool cable while the master robot is not holding the tool. Fig. \ref{fig:cablehold} exemplifies two snapshots grabbed from an execution record where the assistant robot holds the cable, allowing the master robot to release and regrasp the tool. 

\begin{figure}[!htpb]
  \begin{center}
  \includegraphics[width=.95\linewidth]{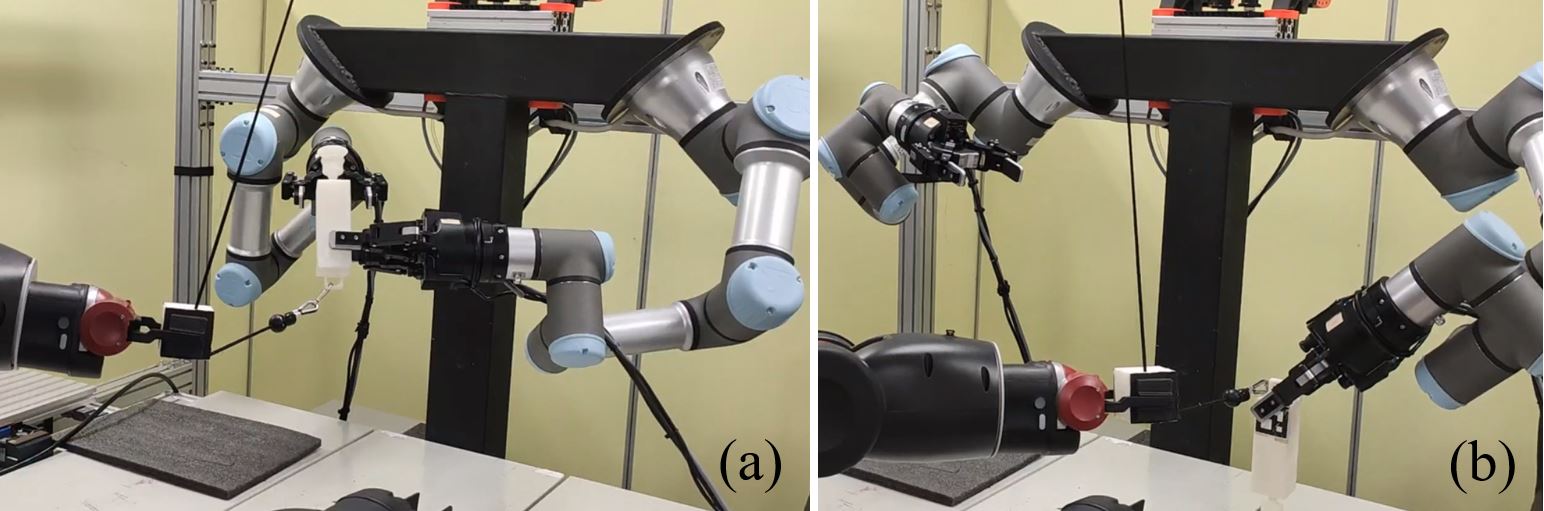}
  \caption{The assistant robot helps the master robot by controlling the cable shape during the whole manipulation sequence. The master robot performs a handover (shown in the center picture) to place the tool in the desired goal pose. The freedom given by the assistant robot lets the master robot use both hands to handle the tool and cover its whole workspace.} 
  \label{fig:toolbalancerms}
  \end{center}
\end{figure}

For handover, executions require several tool rotations and reposes in order for the master robot to pass the tool from one of its hands to the other hand. With our planner, the assistant robot can follow the movements of the tool, avoiding entanglements while the master robot performs handovers and the rest of the planned motion sequence. Fig. \ref{fig:toolbalancerms} showcases the assisted handover motions that follow the master maneuvering. 

\subsubsection{Results using the motorized pulley} 
\begin{figure*}[!htpb]
  \begin{center}
  \includegraphics[width=.95\textwidth]{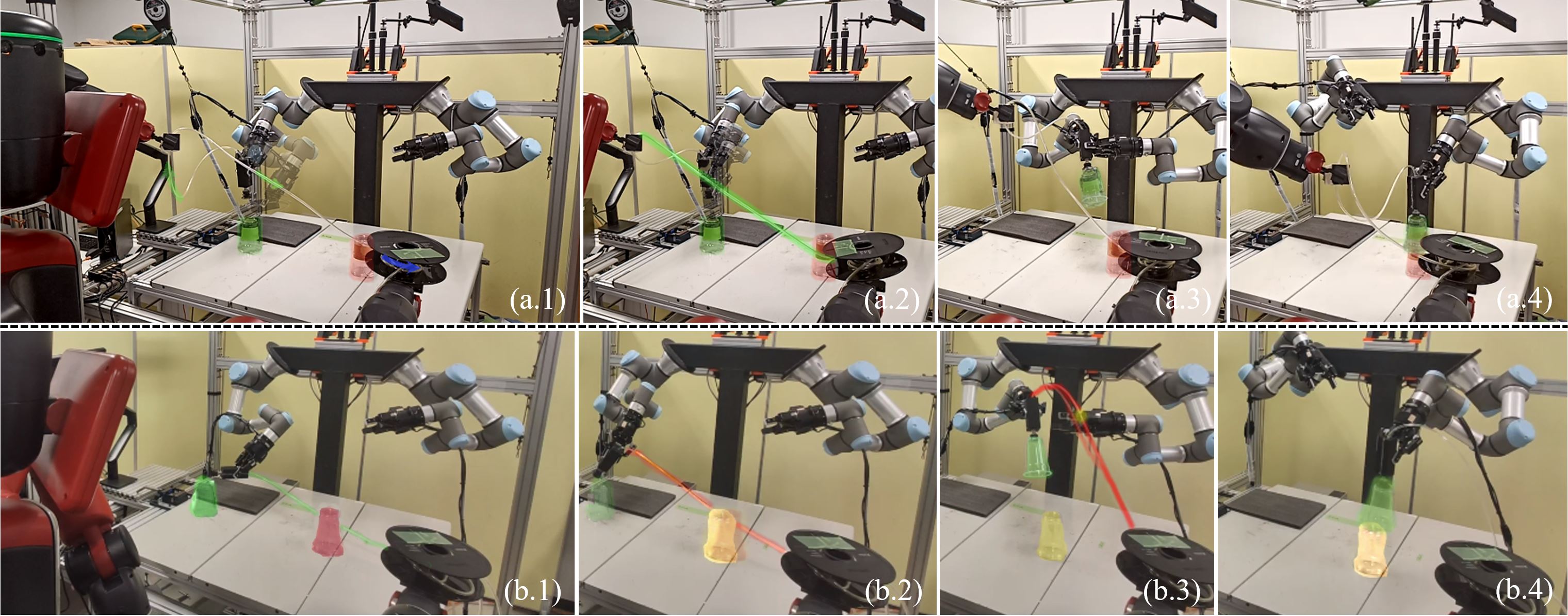}
  \caption{Planning to use a suction cup to pick up a plastic object (green) and place it down over another (red). (a.1-4) A successful result where the assistant robot uses the motorized pulley to straighten the tool cable and avoid entanglement and collision. (b.1-4) A TMS-only execution sequence where the unassisted master robot fails to avoid a cable collision -- the cable moves the goal object out of place at (yellow, b.2) and the robot collides and almost grasps the cable at (b.3). The master robot failed to place the object right over the goal.} 
  \label{fig:pulleyexp}
  \end{center}
\end{figure*}

In these experiments, one of the assistant robot hand holds the motorized pulley and use it to help handle the cables. The holding position is considered as the $h$ point used in equations \eqref{eq:slider to balancer segment}. The experiments particularly require the robot to use a suction cup to grab an object and then place it in the desired goal position. Like the real-world experiments using a tool balancer, the planner must find an AMS that avoids collisions between the cables and other objects in the robot workspace. Meanwhile, the pulley motor control signals must be planned together with tool placements and handovers to adjust the cable for completing the task. Fig. \ref{fig:pulleyexp}(a) exemplifies some snapshots grabbed from a successful execution record of our proposed planner considering pulley motor control. For comparison, a TMS-only execution sequence is shown below in  Fig. \ref{fig:pulleyexp}(b). 

\section{Conclusions and Future Work}

In this paper, we presented a planner for tethered tool manipulation. The planner enables a pair of dual-armed robots to collaboratively perform tethered tool manipulation tasks while avoiding cable collisions and entanglements. The planner is realized hierarchically by considering a TMS, a slider trajectory that follows the TMS, and an AMS that afford the cable trajectory. Besides, policies are included to allow assistant robots to strategically grab and suppress the tool cable tension for releasing and regrasp tools. Experimental results show a clear improvement in the manipulation task execution success. The planner allows the master robot to execute both handovers and object placements, which contrasts with previous methods that forbid the robot from executing both actions in the same motion sequence.

This paper assumes that the base position of the assistant robot to be static. It was fixed in front of the master robot for all experiments. Future research will be aimed at mobile assistant robots, and plan the base movement considering balance requirements to perform the AMS.
 
\bibliographystyle{IEEEtran}
\bibliography{paperDaniel}

\end{document}